\documentclass{cta-author}

{}
{}
{}

\usepackage{graphicx}
\usepackage{amsmath}
\usepackage{booktabs}
\usepackage{multirow}
\usepackage{pythonhighlight}
\usepackage{algorithm}
\usepackage{listings}

\usepackage{hyperref}

\begin{document}

\supertitle{CAT: Learning to Collaborate Channel and Spatial Attention from Multi-Information Fusion}

\title{CAT: Learning to Collaborate Channel and Spatial Attention from Multi-Information Fusion}

\author{\au{Zizhang Wu$^{1\corr}$, \au{Man Wang$^{1}$}, \au{Weiwei Sun$^{1}$}}, \au{Yuchen Li$^{1}$}, \au{Tianhao Xu$^{1,2}$}, \au{Fan Wang$^{1}$}, \au{Keke Huang$^{3}$}}

\address{\add{1}{Zongmu Technology}
\add{2}{Technical University of Braunschweig}
\add{3}{Central South University}
\email{wuzizhang87@gmail.com xutianhao2018@gmail.com}}

% \begin{abstract}
% This should be informative and suitable for direct
% inclusion in abstracting services as a self-contained
% article. It should not exceed 200 words. It should
% summarise the general scope and also state the main results
% obtained, methods used, the value of the work and the
% conclusions drawn. No figure numbers, table numbers,
% references or displayed mathematical expressions should be
% included. The abstract should be included in both the
% Manuscript Central submission step (Step 1) and the
% submitted paper.
% \end{abstract}

\begin{abstract}
Channel and spatial attention mechanism has proven to provide an evident performance boost of deep convolution neural networks (CNNs).
Most existing methods focus on one or run them parallel (series), neglecting the collaboration between the two attentions. 
In order to better establish the feature interaction between the two types of attention, we propose a plug-and-play attention module, which we term “CAT”—activating the Collaboration between spatial and channel Attentions based on learned Traits. 
Specifically, we represent traits as trainable coefficients (i.e., colla-factors) to adaptively combine contributions of different attentions modules to fit different image hierarchies and tasks better. 
Moreover, we propose the global entropy pooling (GEP) apart from global average pooling (GAP) and global maximum pooling (GMP) operators, an effective component in suppressing noise signals by measuring the information disorder of feature maps. 
We introduce a three-way pooling operation into attention modules and apply the adaptive mechanism to fuse their outcomes. Extensive experiments on MS COCO, Pascal-VOC, Cifar-100, and ImageNet show that our CAT outperforms existing state-of-the-art attention mechanisms in object detection, instance segmentation, and image classification. 
The model and code will be released soon.

\end{abstract}

\maketitle

\section{Introduction}
Convolutional neural networks (CNNs) are one of most powerful learning algorithms for vision tasks and have shown exemplary performance in areas like image classification, object detection and semantic segmentation \cite{article1,article2,article5,article6,article7}. The general form of CNNs usually consists of multiple stages dealing with hierarchical features, exploiting spatial or other correlations in data at a multi-level. There are three primary factors during the learning process: sparse interaction, parameter sharing, and equivariant representation.

From the three perspectives above, various adjustments and improvements were performed on CNN to deal with more complex and heterogeneous problems. CNN based methods became extremely prevalent after the great success of AlexNet \cite{Alex}. After that, innovations in CNN components ushered in rapid development. The split, transform and merge became a routine process, allowing the abstraction of features at different spatial scales. Afterwards, the concept was applied to most succeeding methods, including the attention mechanisms.

The performance boost of CNN mainly lies in adjustments on network structure \cite{article8,article9} and introductions of effective normalization and pooling mechanisms \cite{article10,article11, article12}. Attention mechanism works on assigning extra weights to significant factors through applying residual connections on CNN backbones. Recent attention-based networks \cite{article13,article14,article15} have made preferable improvements in tasks of classification, segmentation, and detection, etc. 

\begin{figure}[!ht]
    \centering
    \includegraphics[width=8.2cm]{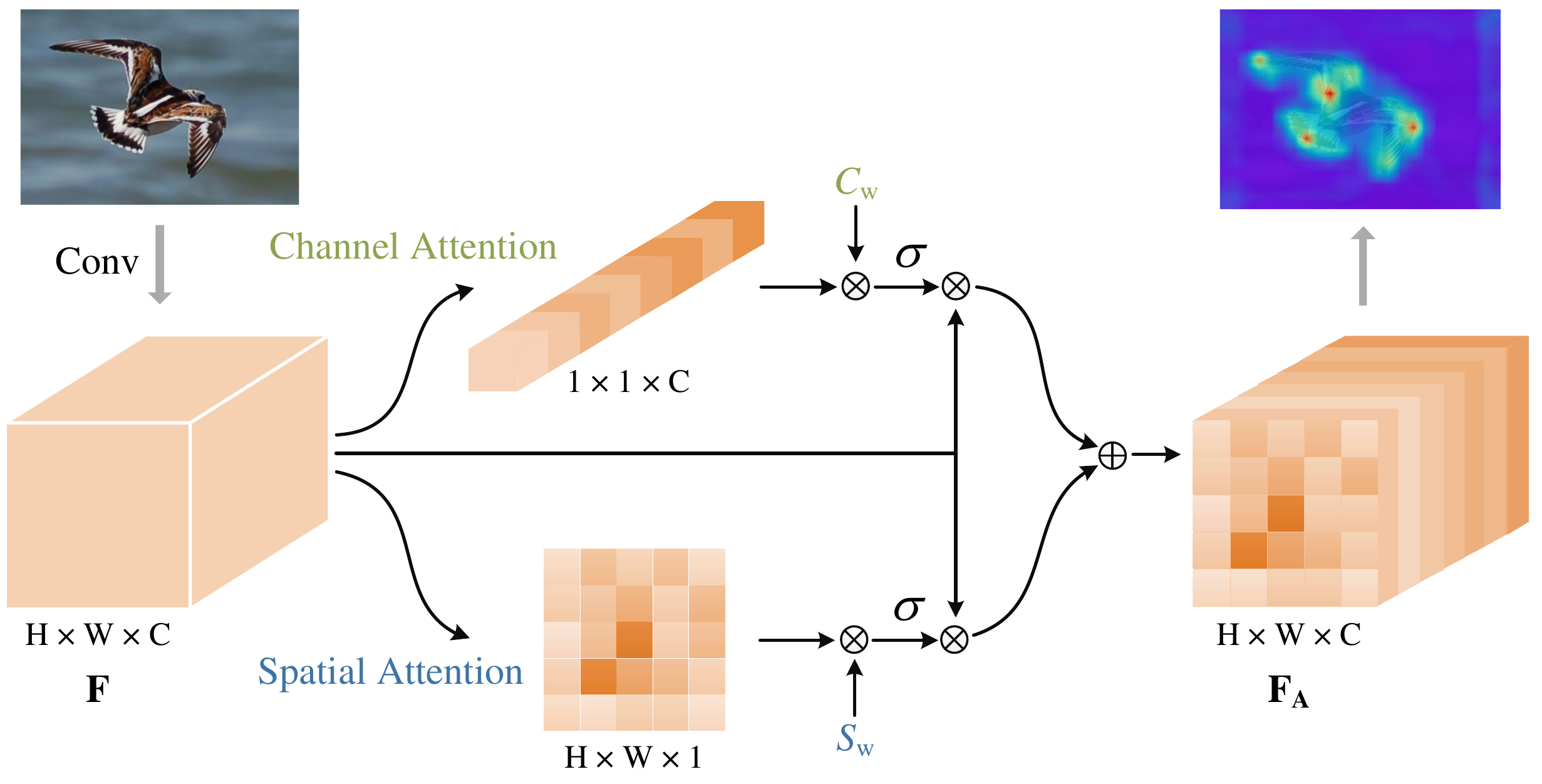}

    \caption{Our CAT framework --
    Channel and spatial attention collaborate via the linear combination, which is controlled by the exterior colla-factors (i.e., $C_\text{w}$ and $S_\text{w}$). 
    We set the colla-factors as trainable parameters in network. 
    Additionally,to generate attention, we fuse information collected from three pooling methods -- i.e., global max pooling (GMP), global average pooling (GAP) and our proposed  global entropy pooling (GEP). 
    $\otimes$ and $\oplus$ denote element-wise multiplication and element-wise addition. $\sigma$ is a sigmoid function. Best view in color and zoom in.}

    \label{fig:Figure01}
\end{figure}

Squeeze-and-Excitation Networks (SENet) \cite{article13}, the most representative attention-based network, designs a squeeze-and-excitation module that extracts channel-wise weights through applying GAP operators and assigns them to each spatial plane of a feature map. 
Such architecture brings notable performance gain compared to baselines and costs little extra computational capacity. 
ECANet (Efficient Channel Attention) \cite{article15} designs a 1-D convolution kernel to conduct the interaction among channels so that it further reduces the computational complexity. Later researches \cite{article16,article17,article18,article19} work on developing techniques of squeeze and excitation only achieve limited progress due to the lack of information about spatial-wise interaction \cite{article20,article21,article22}. To tackle this problem, \cite{article25,article26,article27} suggest architectures to take both spatial and channel interactions into consideration. Particularly, Convolutional Block Attention Module (CBAM) \cite{article27} achieves notable improvements than SENet through designing a spatial-wise attention module and sequentially connect it with the channel-wise attention module.

Most previous attention-based methods only focus on the channel or spatial attention, or embed them through direct addition or concatenation \cite{article24,article25,article27}. Given such a situation, an adaptive mechanism could benefit the fusion of different attentions more feasibly. In other words, our framework Collaborates channel and spatial Attention contributions based on their Traits (CAT) in a multi-information-fusion style. We also notice that the widely used GAP \cite{article10} and GMP \cite{article32} operators have disadvantages in capturing rich texture distribution information of large ranges. Moreover, they are probably to cause excessive background noise as the former operator treats the background and object regions equally and the latter one is merely susceptible to extreme features \cite{article37}. Hence, we introduce GEP operators into our network, which measure spatial and channel systems' information disorders by calculating the entropy of input feature maps (Section \ref{section:3}). Our ablation experiments illustrate that placing these attention operators in parallel and fusing their outcomes with the adaptive mechanism makes our framework the best performance in segmentation and detection tasks, etc.

The overview of the CAT framework is specifically present in Fig. \ref{fig:Figure01}. We disentangle channel and spatial attention modules, and assign two exterior colla-factors $C_{\text{w}}$ and $S_{\text{w}}$ to them to attain the coordination and cooperation of the two modules. It is noteworthy that we can embed the CAT in all appropriate positions in networks, such as the residual operation of each block in ResNet. In each attention module, we use the GEP, GMP, and GAP to extract three raw attention maps from input features $F$ for channel and spatial in $\mathbb{R}^{1\times1\times \text{C}}$ and $\mathbb{R}^{\text{H}\times \text{W}\times 1}$. Inside the channel attention module, to effectively combine the outputs of the above operators, we adopt an element-wise addition to equilibrate their effects and design a shared multilayer perception (MLP) structure to capture their cross-channel interactions. On the other hand, inside the spatial attention module, we construct a system consisting of a dynamic and linearly weighted fusion operator and a $7\times7$ convolution to capture the spatial neighbourhood information.

We verify the performance of CAT in object detection, instance segmentation, and image classification on Pascal-VOC, MS COCO, and Cifar-100 datasets, respectively. Extensive experiments show that the proposed CAT outperforms state-of-the-art attention mechanisms such as SENet \cite{article13}, CBAM \cite{article27} and ECANet \cite{article15}.

Our contributions can be summarized as follows: (1) We propose a collaborative attention framework by dynamically learning the interaction between channel and spatial attention modules so that our framework is adaptive to different embedded image hierarchies and tasks; (2) We propose the GEP attention operator and design an adaptive mechanism to capture the inherent collaboration relationship of different attention operators (GEP, GAP, and GMP), which increases the texture sensitivity of our framework while brings negligible extra parameters; (3) We show our network outperforms previous attention-based architectures with no large extra computational complexity required. Furthermore, it reveals the potential of elaborately designed attention extraction structures that selectively emphasise interest factors in a wide range of computer vision tasks.

\section{Related Work}

\paragraph{Attention Mechanism} Many recent researches focus on applying attention mechanisms in a series of computer vision tasks \cite{article13, article32, article15}. 
The most representative attention-based network, SENet \cite{article13}, employs GAP to obtain weights of attention for spatial planes and conducts cross-channel interaction via using fully-connection layers. 
Later arts like GENet (Gather-excite) \cite{article16}, SGENet (Spatial group-wise enhance) \cite{article31}, GCNet \cite{article32}, and SKNet \cite{article33} work on improving SENet (Selective kernel network) \cite{article13} in its mechanism of extracting attention weights and its fusing method of combining attention module outputs and feature maps.
However, the employment of fully-connection layers in these works for conducting information interaction among channels leads to great computational complexity. 
To tackle this problem, ECANet \cite{article15} employs a 1-D convolution kernel instead that boosts the training and implementation speed with competitive network performance. 

Other researches start to present attention modules in the spatial aspect \cite{article20,article21,article22,article34} to further improve the performance of neural networks.
Spatial Transformer Network \cite{article20} pays additional attention to regions of interest and transforms them into expected postures to facilitate the learning of backbone and output layers. 
\cite{article34} uses the summation operator to replace GAP for alleviating the loss of information caused in person-type-object re-identification. 
For capturing long-range dependencies, \cite{article14} proposes a non-local method to measure relationships of all possible couples for whole sets of pixels in feature maps.
To further combine channel and spatial attention contributions, CBAM \cite{article27} sequentially introduces both of them into attention modules in which GAP and GMP act simultaneously to extract attention weights. 
Networks of \cite{article24,article26} adopt a similar non-local way to deal with spatial attention and combine the spatial and channel outcomes in parallel structure for segmentation tasks. 

Along with the CBAM, Bottleneck Attention Module (BAM \cite{BAM}) also infers the attention map along two separate pathways, channel and spatial, but constructs hierarchical attention at bottlenecks.
Recently, FacNet \cite{fcanet} consider the attention mechanism by regarding the channel representation problem as a compression process using frequency analysis, which proved that GAP is a special case of the feature decomposition in the frequency domain. 
CAEM \cite{CAEM} proposed an attention mechanism by embedding positional information into channel attention, the coordinate attention.
There are also works in low-level vision fields that explored the attention mechanism. CycleISP \cite{lowlevel2} and MIRNet \cite{lowlevel1} achieved better results with attention mechanisms in RAW and sRGB image denoising, color matching, image restoration and enhancement tasks.
Despite the impact of above methods on CNNs, they neglect the important difference between channel-wise and spatial-wise attention contributions, which is important when dealing with various image hierarchies and tasks. 
We thus propose a framework to dynamically learn collaborating exterior and interior interactions that are between and within different attention modules, respectively.

\begin{figure*}[!ht]
        \centering
		\begin{minipage}[a]{0.8\linewidth}
			\centering
			\centerline{\includegraphics[width=14cm]{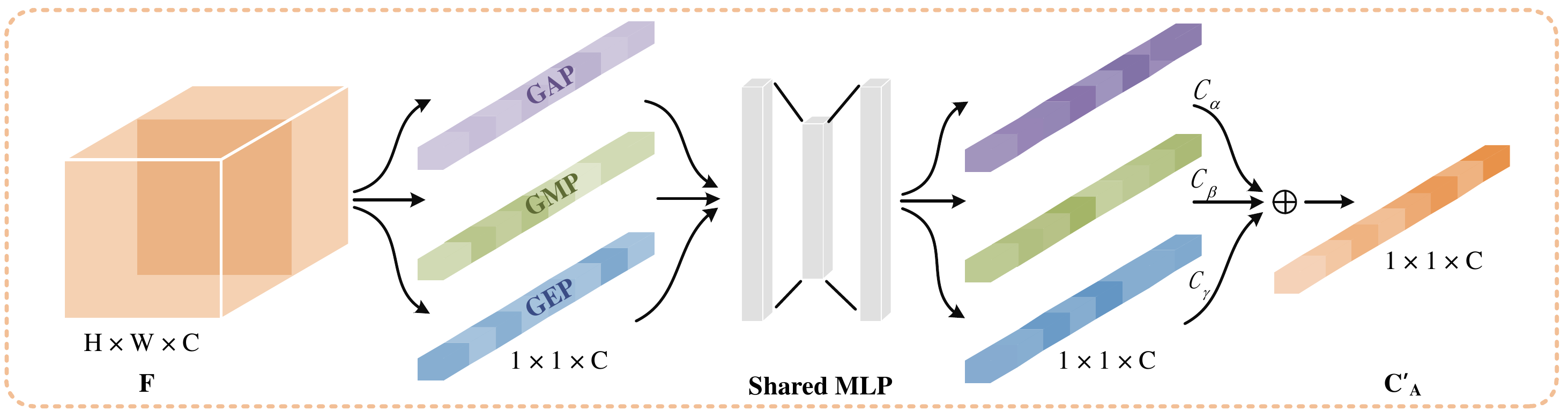}}
			\centerline{\small (a)}
		\end{minipage}
		
		\begin{minipage}{0.8\linewidth}
			\centering
			\centerline{\includegraphics[width=14cm]{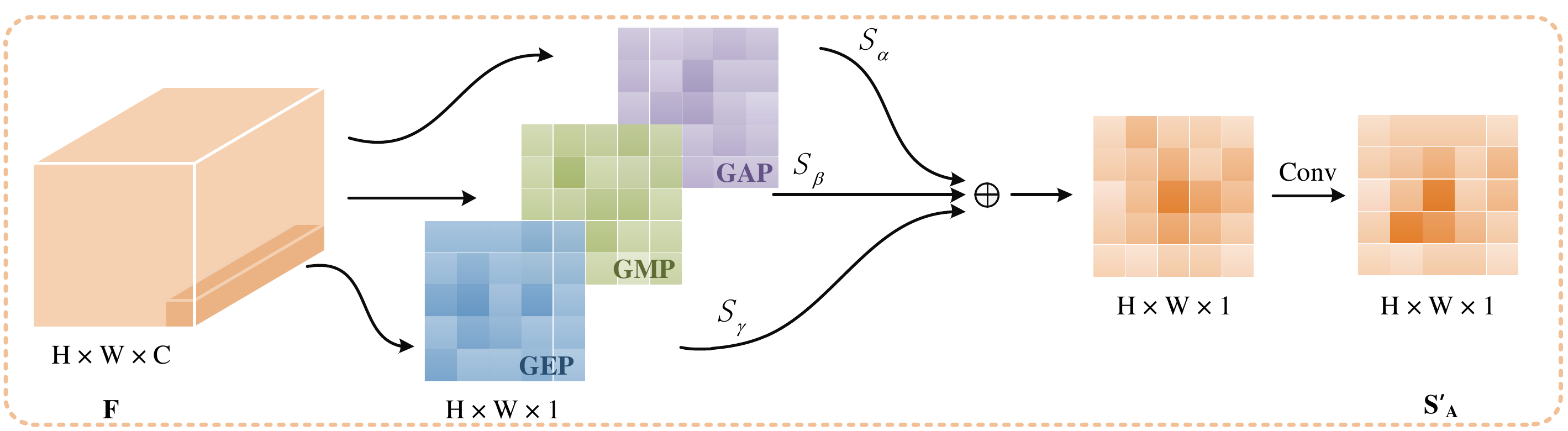}}
			\centerline{\small (b)}
			
		\end{minipage}
		
        \vspace{0.5em}
        \caption{Illustrations of the attention modules: (a) Channel attention module. We employ channel GAP, GMP, and GEP to extract raw attention maps. A shared MLP and an element-wise addition are used to combine the outputs. Three interior colla-factors $C_{\alpha }$, $C_{\beta }$, and $C_{\gamma }$ are respectively assigned to GAP, GMP, and GEP to collaborate with them jointly dynamically. (b) Spatial attention module. It gathers outputs of spatial GAP, GMP, and GEP attention maps. Three interior colla-factors $S_\alpha$, $S_\beta$, and $S_\gamma$ are respectively assigned to GAP, GMP, and GEP to collaborate with them jointly dynamically. Best view in color and zoom in.}
        \label{fig:Figure02}
	\end{figure*}
	
\paragraph{Information Entropy} The objective of measuring information entropy is to count the information uncertainty of a system \cite{article28,article29,article30}. 
We expect a system to have high uncertainty if its predictions are of large entropy and vice versa  \cite{article37}. 
The Entropy Guided Adversarial (EGA) model proposed in  \cite{article37} introduces an entropy loss function for guiding the CNN to make pixel-level detections of objects. 
Some other methods also employ entropies for information pruning \cite{article28,article29,article38}. 
\cite{article28} combines impacts of kernel sparsity and entropy to quantify the importance of a feature map for the task of model compression.
\cite{article29} also constructs a quantitative model by defining a weighted entropy to comprehensively measure the importance and frequency of filters.

There are also successful applications of entropy in semi-supervised and inter-domain adaptive learnings \cite{article30,article39,article40,article41}. 
For example, \cite{article30} employs the minimax entropy (MME) method to train its classifier through reducing differences among distributions and learning discriminative features in tasks. 
The utilization of entropy minimization \cite{article39} can train the classifier from unmarked or partially marked data. 
Moreover, \cite{IEpool} proposed information entropy based feature pooling but only for classic CNNs.
Inspired by above applications of entropy, we design and introduce a global entropy pooling (GEP) operator for our attention-based framework.
The proposed GEP is able to complement GAP and GMP for restraining the effect of irrelevant background noises during pooling. 
These pooling operators help to capture attention maps in three different formats and information extraction views to improve attention modules’ sensitivity and robustness.

\section{Proposed Method}
\label{section:3}

In this section, we will introduce the proposed CAT framework in detail which focuses on dynamically learning the collaborative relationship between spatial and channel attention modules. In each attention module, the GEP complements GMP and GAP to extract attention weights. We hence apply the adaptive mechanism that firstly captures the collaborative relationship of above pooling operators based on their traits and then fuses spatial and channel attention modules for measuring their interactions.

\subsection{Overview} 

The overall structure of the proposed CAT is shown in Fig. \ref{fig:Figure01}. 
We linearly combine the collaborative relationship between spatial and channel attention modules since they are sensitive to different network depths and tasks.
We introduce exterior colla-factors $C_{\text{w}}$ and $S_{\text{w}}$ for channel and spatial attention modules and multiply them with corresponding attention weights.
We then multiply the modified weights with input feature map $F$ to generate two attention maps. 
Eventually, through applying element-wise addition operation, we can obtain the final feature map $F_{\text{A}}$.
\setlength\abovedisplayskip{5pt}
\setlength\belowdisplayskip{5pt}
\begin{equation}
F_{\text{A}}= (F\otimes \sigma(C'_{\text{A}} C_{\text{w}})) + (F\otimes  \sigma(S'_{\text{A}}S_{\text{w}}))
\label{Eq(7)}
\end{equation}
\noindent
where $C'_{\text{A}}$ and $S'_{\text{A}}$ represent raw attention maps of the channel and spatial dimension. The framework initializes learnable $C_{\text{w}}$ and $S_{\text{w}}$ as 0 and applies a softmax function to normalize them. $\otimes$ denotes an element-wise multiplication. 

\noindent

\subsection{Channel attention module}

$F\in \mathbb{R}^{H\times W\times C}$ is an input feature map, where $W$, $H$, and $C$ are width, height, and channel dimensions. As shown in Fig. \ref{fig:Figure02} (a), for channel attention module, we adopt the GAP, GMP, and GEP to extract three-dimensional attention weights whose dimensions are $\mathbb{R}^{1\times1\times C}$ as follows:
\setlength\abovedisplayskip{5pt}
\setlength\belowdisplayskip{5pt}
\begin{equation}
\centering
\begin{split}
& C_{\text{Avg}}; C_{\text{Max}}; C_{\text{Ent}}= MLP(C_{\text{Avg}}'; C_{\text{Max}}'; C_{\text{Ent}}')
\label{Eq(1)}
\end{split}
\end{equation}

\noindent
where $C_{\text{Avg}}'$, $C_{\text{Max}}'$, and $C_{\text{Ent}}'$ denote the GAP, GMP, and GEP attention maps. MLP is a parameter shared network with two fully-connected (FC) layers and an activation layer. We use the first FC to reduce channel dimension into $\mathbb{R}^{1\times1\times \frac{C}{r}}$, where $r$=16 is the reduction ratio. The following ReLU function activates the former output. Hence, we adopt the last FC to introduce non-linearity and extend channel dimension into $\mathbb{R}^{1\times1\times C}$. 

In order to pay more attention to channels with rich textures, we employ GEP to extract the corresponding attention weight of each channel. Here the GEP is defined as:
\setlength\abovedisplayskip{5pt}
\setlength\belowdisplayskip{5pt}
\begin{equation}
C'_{\text{Ent}}= -\sum_{\text{i=1}}^{H\times W}p_{\text{i}}logp_{\text{i}}
\label{Eq(4)}
\end{equation}

\noindent
where $p_{\text{i}}= \text{softmax}(F^{1\times 1\times C}_{\text{i}})$ is the probability of location $i$ in each feature map. We normalize $C'_{\text{Ent}}$ in the interval [-1, 1] before inputting it into MLP through the function:

\setlength\abovedisplayskip{5pt}
\setlength\belowdisplayskip{5pt}
\begin{equation}
\centering
\chi_{\text{i}}\rightarrow \frac{\chi_{\text{i}}-\chi_{\text{min}}}{\chi_{\text{max}}-\chi_{\text{min}}}
\label{Eq(2)}
\end{equation}
where $\chi_{\text{i}}$, $\chi_{\text{min}}$, and $\chi_{\text{max}}$ are for the operation on the $C'_{\text{Ent}}$.

As shown in Fig. \ref{fig:Figure02} (a), the element-wise addition can obtain the channel attention map $C'_{\text{A}}$. To aggregate the above GAP, GMP, and GEP attention operators adaptively, the interior colla-factors $C_{\alpha }$, $C_{\beta }$, and $C_{\gamma }$ collaborate with the operators as follows:
\setlength\abovedisplayskip{5pt}
\setlength\belowdisplayskip{5pt}
\begin{equation}
\centering
C'_{\text{A}}= C_{\text{Avg}}C_{\alpha }+ C_{\text{Max}}C_{\beta }+ C_{\text{Ent}}C_{\gamma }
\label{Eq(3)}
\end{equation}
\noindent
where $C_{\alpha }$, $C_{\beta }$, and $C_{\gamma }$ are initialized as 0 and learnable through training.
GAP and GMP retain the global average information and the most distinctive information of the input $F$. Extra noises are easy to disturb the distinctive information from GMP. Therefore, we employ a 1-D Gaussian low-pass filter ($k\times1$) before all GMP operations to alleviate the inference that causes adverse effects for obtaining more effective maximum response characteristics. Furthermore, $k$ is set to 5 in this paper after comparisons in experiments.

\subsection{Spatial attention module} 

As shown in Fig. \ref{fig:Figure02} (b), similar to the channel attention module, the spatial attention module also uses the GAP, GMP, and GEP to obtain three attention maps in $\mathbb{R}^{H\times W\times 1}$.  We encode channel information at each pixel over all spatial locations and outputs one feature map.

In spatial attention module, the GEP is defined as:
\setlength\abovedisplayskip{5pt}
\setlength\belowdisplayskip{5pt}
\begin{equation}
S_{\text{Ent}}= -\sum_{\text{j=1}}^{C}p_{\text{j}}logp_{\text{j}}
\label{Eq(5)}
\end{equation}
\noindent
$p_{\text{j}}= \text{softmax}(F^{H\times W\times 1}_{\text{j}})$ is the probability for a channel.

To aggregate the above GAP, GMP, and GEP pooling operators adaptively, we introduce interior colla-factors $S_{\alpha }$, $S_{\beta }$, and $S_{\gamma }$ to collaborate with the operators. We then multiply them with corresponding feature maps as follows:
\setlength\abovedisplayskip{5pt}
\setlength\belowdisplayskip{5pt}
\begin{equation}
S'_{\text{A}}= \sigma(\text{Conv}^{7\times7} (-S_{\text{Avg}}S_{\alpha } + S_{\text{Max}}S_{\beta } + S_{\text{Ent}}S_{\gamma }))
\label{Eq(6)}
\end{equation}

\noindent
where $S_{\text{Avg}}$, $S_{\text{Max}}$, and $S_{\text{Ent}}$ represent the global average, global maximum feature, and information attention maps. $S_{\alpha}$, $S_{\beta}$, and $S_{\gamma}$ are initialized as 0 and learnable through training. $\text{Conv}^{7\times7}$ is a $7\times 7$ convolution kernel and $\sigma(.)$ is a sigmoid function. Moreover, the same as what we do in channel attention module, we normalize $S_{\text{Ent}}$ to $[-1,1]$ . 

\begin{figure}[!b]
    \centering
    \includegraphics[width=0.45\textwidth]{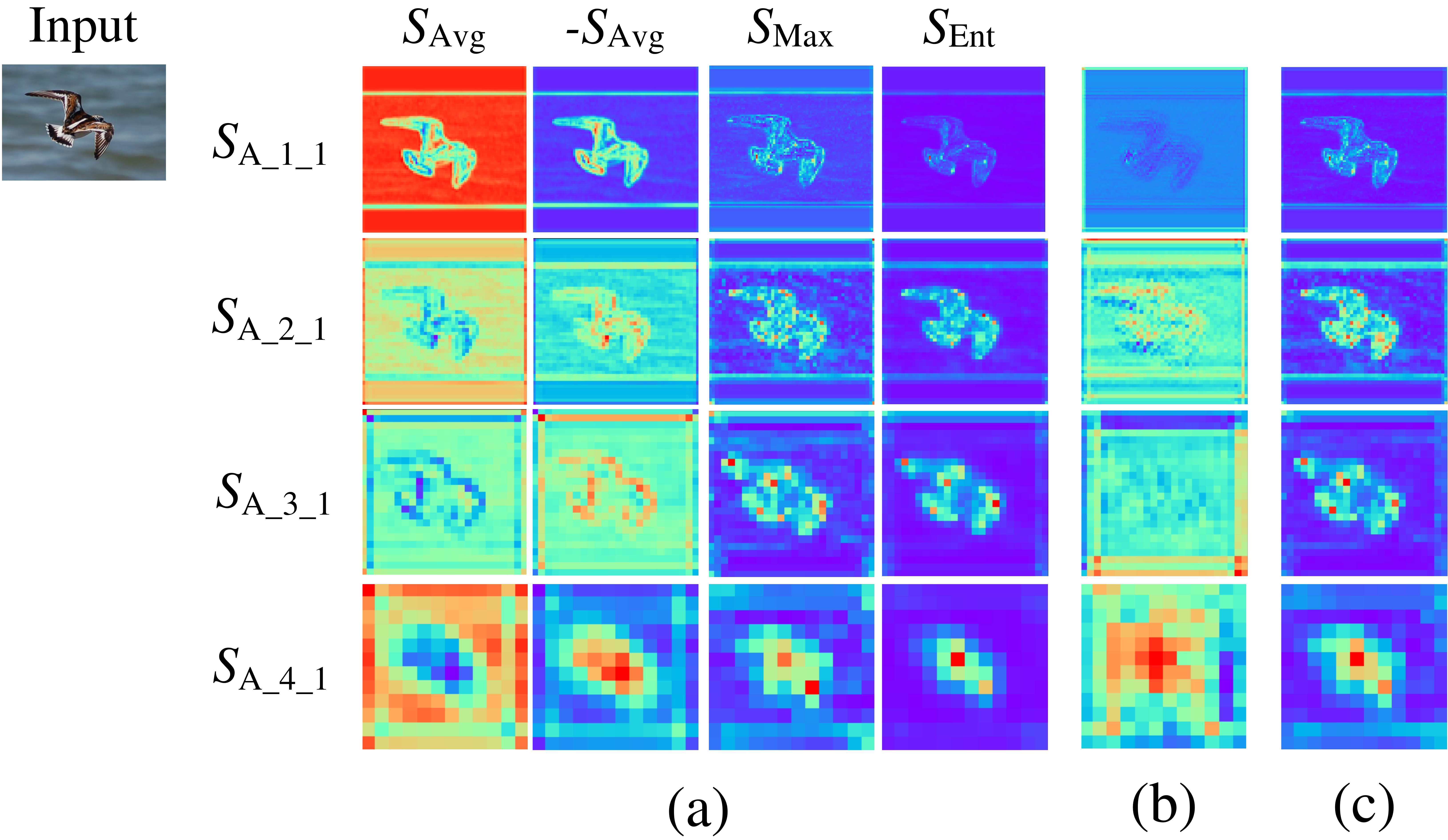}
%    \captionsetup{font={small}}
    %\caption{Attention map visualization.}
    \caption{Visualization of spatial attention maps with different fusion methods. (a) Raw attention maps in first block of each stage in ResNet50. Each column corresponds to the attention maps of  GAP ($S_{\text{Avg}}$, $-S_{\text{Avg}}$), GMP ($S_{\text{Max}}$), and GEP ($S_{\text{Ent}}$). (b) Concatenated fusion attention maps. (c)  Fusion attention maps from weighted summation. Best view in color and zoom in.}
    \label{fig:Figure03}
\end{figure}

In our framework, we carry out a negative operation for $S_{\text{Avg}}$ and apply an element-wise addition for feature fusion. As shown in Fig. \ref{fig:Figure03} (a), the contributions of $S_{\text{Avg}}$ is different from that of $S_{\text{Max}}$ and $S_{\text{Ent}}$. In $S_{\text{Avg}}$, the response intensity of the background is significantly stronger than that of the object, while the opposite phenomenon appears for $S_{\text{Max}}$ and $S_{\text{Ent}}$. Conducting the negative operation on $S_{\text{Avg}}$ ($-S_{\text{Avg}}$) ensures that responses of different attention feature maps have the same characteristics. Using a 2-D Gaussian low-pass filter ($k\times k$) before performing the GMP is also useful to alleviate the adverse effects caused by noises to obtain $S_{\text{Max}}$. We set $k$ to be 5 as well.

We have verified two methods for fusing these attention maps. As shown in Fig. \ref{fig:Figure03} (c), element-wise addition is more advantageous as it brings a dynamic fusing method to adapt contributions of different attention maps. Moreover, we employ a $7\times7$ convolution to capture spatial neighborhood features. In the end, we employ a sigmoid function $\sigma(.)$ on the spatial attention map. The pseudo-code of CAT is shown in Algorithm \ref{alg:code}.

% \caption{Pseudo-code of the proposed CAT module.}

\begin{algorithm}[hb!]

\caption{CAT, PyTorch-like}
\label{alg:code}
\definecolor{codeblue}{rgb}{0.25,0.5,0.5}
\definecolor{codekw}{rgb}{0.85, 0.18, 0.50}
\lstset{
  backgroundcolor=\color{white},
  basicstyle=\fontsize{7.5pt}{7.5pt}\ttfamily\selectfont,
  columns=fullflexible,
  breaklines=true,
  captionpos=b,
  commentstyle=\fontsize{7.5pt}{7.5pt}\color{codeblue},
  keywordstyle=\fontsize{7.5pt}{7.5pt}\color{codekw},
}

\begin{lstlisting}[language=python]
# feat: Input feature maps, BxCxWxH
# w1, w2, w3: Trainable coefficients for three types of channel information. 
# w4, w5, w6: Trainable coefficients for three types of spatial information. 
# w7, w8: Trainable coefficients between channel and spatial scores.

att = CAT(feat)
feat = att * feat

def CAT(feat): # Compute CAT attention
    # Aggregate channel information via three ways.
    channel_avg = pool(feat, dim=(2, 3), method="Avg", keepdim=True) # BxCx1x1
    channel_max = pool(feat, dim=(2, 3), method="Max", keepdim=True) # BxCx1x1
    channel_entropy = pool(feat, dim=(2, 3), method="Entropy", keepdim=True)# BxCx1x1 
    
    # Linear combination of channel information. 
    channel_score = w1* channel_avg + w2 * channel_max+ w3 * channel_entropy
    channel_score = Sigmoid(channel_score)

    # Aggregate spatial information via three ways.
    spatial_avg = pool(feat, dim=(1), method="Avg", keepdim=True) # Bx1xWxH
    spatial_max = pool(feat, dim=(1), method="Max", keepdim=True) # Bx1xWxH 
    spatial_entropy = pool(feat, dim=(1), method="Entropy", keepdim=True) # Bx1xWxH  
    
    # Linear combination of channel information. 
    spatial_score = -w4 * spatial_avg + w5 * spatial_max+ w6 * spatial_entropy
    
    spatial_score = conv(spatial_score)
    spatial_score = Sigmoid(spatial_score)
    
    return  w7 * channel_score + w8 * spatial_score
\end{lstlisting}

\end{algorithm}

\begin{figure*}[!ht]
    \centering
    \includegraphics[width=1\textwidth]{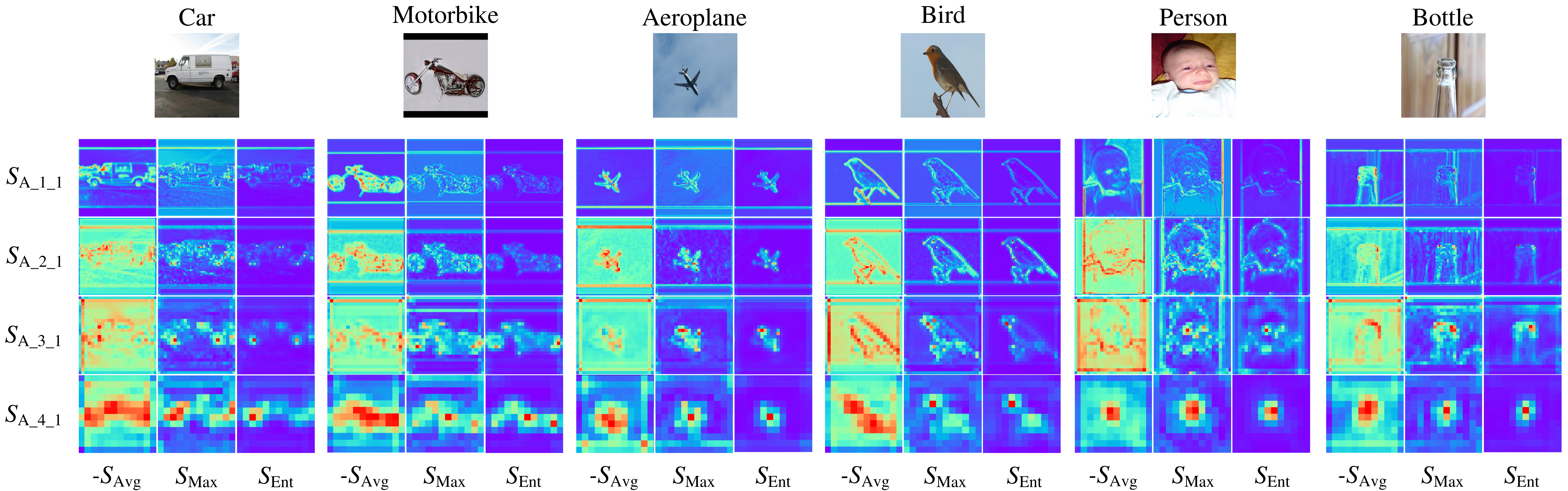}
%    \captionsetup{font={small}}
    %\caption{Attention map visualization.}
    \caption{The visualization of spatial attention maps. From top to bottom, these are the  spatial attention maps for the first block in Resnet50 four stages. Each column of each subgragh corresponds to the attention maps of GAP ($-S_{\text{Avg}}$), GMP ($S_{\text{Max}}$), and GEP ($S_{\text{Ent}}$). It is obvious that the background of the $S_{\text{Ent}}$ is cleaner than the $-S_{\text{Avg}}$ and $S_{\text{Max}}$, and has stronger response to the key texture features such as the tire of a vehicle, the beak of a bird, the wing of an aeroplane, and the bottle mouth. Best view in color and zoom in.}
    \label{fig:Figure04}
\end{figure*}

As shown in Fig. \ref{fig:Figure04}, it is obvious that $-S_{\text{Avg}}$ pays more attention to the overall information of the input, and considers the background and foreground equivalence at the same time. If the response intensity of a small object is high and the features obtained after the GAP are very small, the features of the small object may not be well preserved. $S_{\text{max}}$ extracts the global maximum feature of the input. It can retain the features of strong response such as edges or corners, but it is susceptible to extreme values. 

In $S_{\text{Ent}}$, although contour features of the object are not comprehensive as $S_{\text{max}}$, it has a stronger response to key texture features such as the tire of a vehicle, the beak of a bird, the wing of an aeroplane, and the bottle mouth, etc. Consequently, the background of $S_{\text{Ent}}$ is cleaner than the other two attention maps and is less susceptible to background noises.

\section{Experiments}

\begin{table}[!b]
\small
\setlength\tabcolsep{11pt}
\textwidth 8.7cm
\centering

\scalebox{0.9}{\begin{tabular}{clccc}

\hline
            Detectors                  & Methods   & \multicolumn{1}{c}{$AP$}    & \multicolumn{1}{c}{$AP_{50}$}   &\multicolumn{1}{c}{$AP_{75}$}   \\ \hline
\multirow{10}*{Faster-RCNN} & ResNet50  & 50.47 & 79.25 & 54.14 \\
                            & + SENet    & 50.28 & 79.22 & 54.22 \\
                            & + CBAM      & \underline{50.72} & 79.52 & \underline{54.32} \\
                            & + ECANet   & 50.58 & \underline{79.83} & 54.26 \\
                              & + CAT (Ours)   & \textbf{52.54} & \textbf{80.96} & \textbf{57.46} \\ \cmidrule{2-5}
                              & ResNet101  &  53.10       &   80.98      &    58.45     \\
                              & + SENet    &  \underline{53.26}       &    81.00     &   \underline{58.96}      \\
                              & + CBAM      &    53.19     &   81.03      &   58.69      \\
                              & + ECANet   &     53.16    &   \underline{81.28}      &   58.06      \\
                              & + CAT (Ours)   & \textbf{54.33}   &  \textbf{81.94} &   \textbf{60.11} \\ \hline
\multirow{10}{*}{RetinaNet}   & ResNet50  & 53.11 & 77.92 & 57.03 \\
                              & + SENet    & 53.10 & 78.54 & 57.00 \\
                              & + CBAM      & \underline{53.43} & 77.17 & \underline{57.81} \\
                              & + ECANet   & 53.30 & \underline{79.01} & 57.44 \\
                              & + CAT (Ours)   & \textbf{54.15} & \textbf{79.93} & \textbf{58.66} \\ \cmidrule{2-5} 
                              & ResNet101 &     54.38    &    78.79     &   58.68      \\
                              & + SENet    & 54.94 & 79.57 & 59.36 \\
                              & + CBAM      & 53.85 & 78.53 & 58.44 \\
                              & + ECANet   & \underline{55.12} & \underline{79.70} & \underline{59.57} \\
                              & + CAT (Ours)   & \textbf{55.42} & \textbf{79.94} & \textbf{59.91} \\ \hline
\end{tabular}}
% \captionsetup{font={small}}
%\caption{\textbf{Object detection -- } Our method consistently outperforms other attention mechanisms on Pascal-VOC test 2007. }
\vspace{0.5em}
\caption{Object detection results of different attention methods on Pascal VOC test 2007. }
\label{table01}
\end{table}

% Please add the following required packages to your document preamble:
% \usepackage{multirow}
\begin{table*}[b]
\centering
\setlength\tabcolsep{22pt}
\small
% \captionsetup{font={small}}

%\begin{tabular}{llcccccc}
\scalebox{0.9}{\begin{tabular}{p{1.7cm}<{\centering}lcccccc}
\hline
                             Detectors & Methods  & $AP$    & $AP_{50}$  &  $AP_{75}$  &  $AP_{S}$   &  $AP_{M}$   &  $AP_{L}$   \\ \hline
\multirow{10}{*}{Faster-RCNN} & ResNet50  & 33.27 & 53.58 & 35.26 & 18.04 & 35.71 & 42.56 \\
                              & + SENet    & 33.56 & 53.67 & 36.02 & 18.52 & 36.23 & 42.90 \\
                              & + CBAM     & 33.37 & 53.67 & 35.61 & 19.38 & 35.84 & 43.60 \\
                              & + ECANet   & \underline{34.41} & \underline{54.91} & \underline{37.13} & \underline{20.34} & \underline{37.08} & \underline{43.72} \\
                              & + CAT (Ours)   & \textbf{34.99} & \textbf{55.80} & \textbf{37.69} & \textbf{20.55} & \textbf{37.86} & \textbf{44.86} \\ \cmidrule{2-8} 
                              & ResNet101 &   35.64   & 55.71     & 38.49    & 20.55  & 38.70    & 45.37 \\
                              & + SENet    &    35.70   &   55.97    &   38.54    &   20.54    &   38.99    &    45.61   \\
                              & + CBAM     &  \textbf{36.85}     &   \textbf{57.78}    &  \underline{39.69}     &   \textbf{21.98}    &   \textbf{40.02}    &   \underline{46.85}    \\
                              & + ECANet   &   36.18    &   56.64    &   38.85    &   21.50    &   39.16    &    46.44   \\
                              & + CAT (Ours)   &   \underline{36.78}    &   \underline{57.68}    &   \textbf{39.86}    &   \underline{21.71}    &   \underline{39.97}    &   \textbf{46.99}    \\ \hline
\multirow{10}{*}{RetinaNet}   & ResNet50  & 34.39 & 53.01 & 37.04 & 18.30 & 38.05 & 44.14 \\
                              & + SENet    & 34.50 & 53.07 & 37.01 & 18.90 & 38.08 & 44.29 \\
                              & + CBAM     & 34.38 & 52.95 & 36.60 & 19.60 & 38.21 & 44.22 \\
                              & + ECANet   & \underline{35.23} & \underline{54.13} & \underline{37.70} & \underline{19.78} & \underline{39.27} & \underline{45.29} \\
                              & + CAT (Ours)   & \textbf{36.15} & \textbf{55.38} & \textbf{38.53} & \textbf{21.40} & \textbf{40.12} & \textbf{47.10} \\ \cmidrule{2-8} 
                              & ResNet101 &  35.48     &   53.70    &   37.89    &   20.07    & 39.02      &   45.04    \\
                              & + SENet    &   35.49    &   54.04    &   38.09    &   20.67    &   39.06    &   45.45    \\
                              & + CBAM     &   35.79    &   54.50    &   38.09    &   20.97    &   39.57    &    \underline{46.32}   \\
                              & + ECANet   &   \underline{36.05}    &    \underline{54.92}   &  \underline{38.47}     &  \textbf{21.17}     & \underline{40.08}      &    45.67   \\
                              & + CAT (Ours)  &   \textbf{36.45}    &   \textbf{55.31}    &   \textbf{39.07}    &    \textbf{21.17}   &  \textbf{40.24}    &    \textbf{46.63}   \\ \hline
\end{tabular}}
\vspace{0.5em}
\caption{Object detection results of different attention methods on MS COCO val2017. }\label{table02}
\end{table*}

\begin{figure*} [!ht] 
\begin{minipage}[t]{0.33\linewidth}  
% \centering  
\includegraphics[width=2.4in]{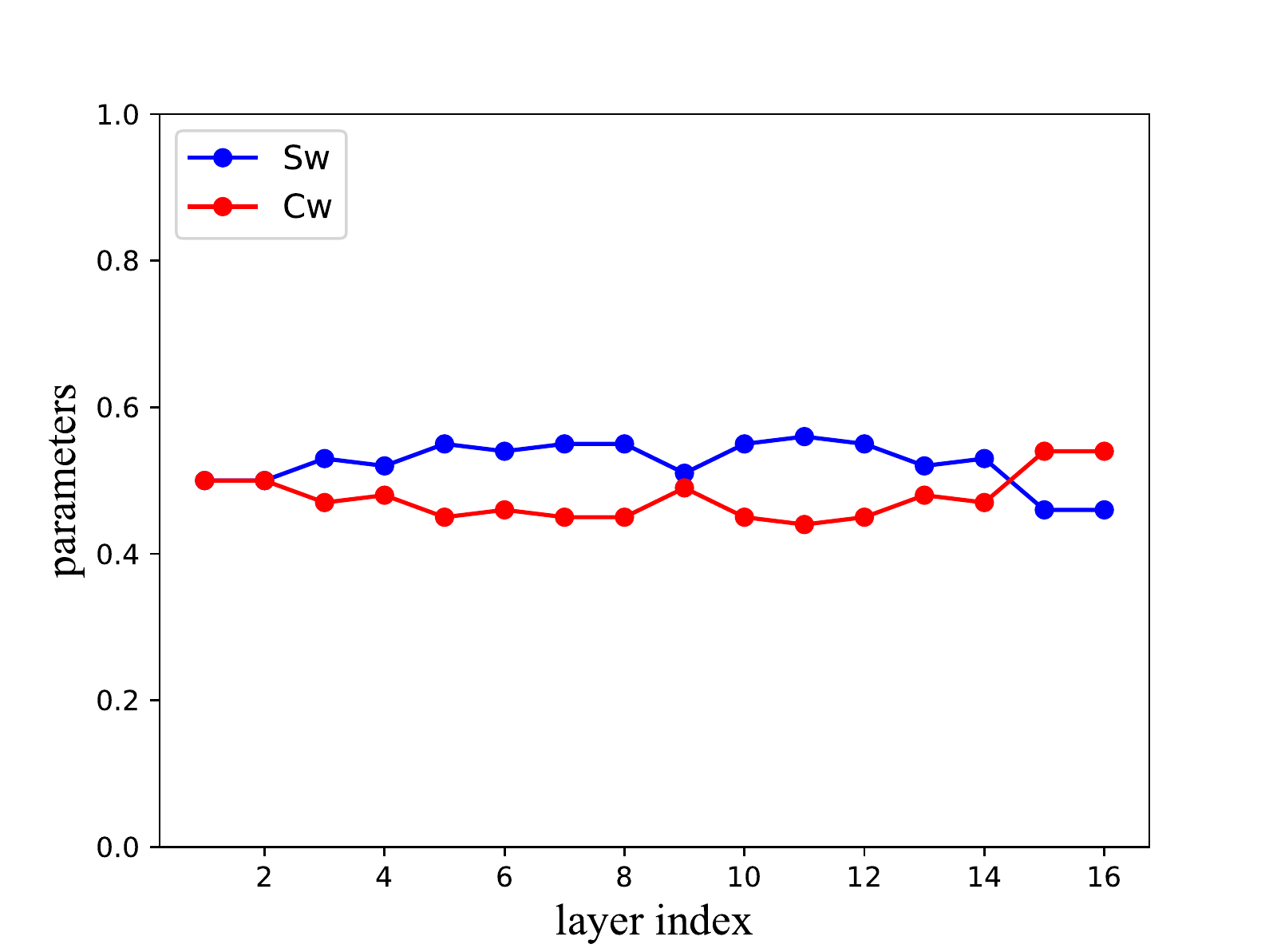}  
\centerline{(a) \small Object Detection}
\end{minipage}%  
\begin{minipage}[t]{0.33\linewidth}  
% \centering  ty
\includegraphics[width=2.4in]{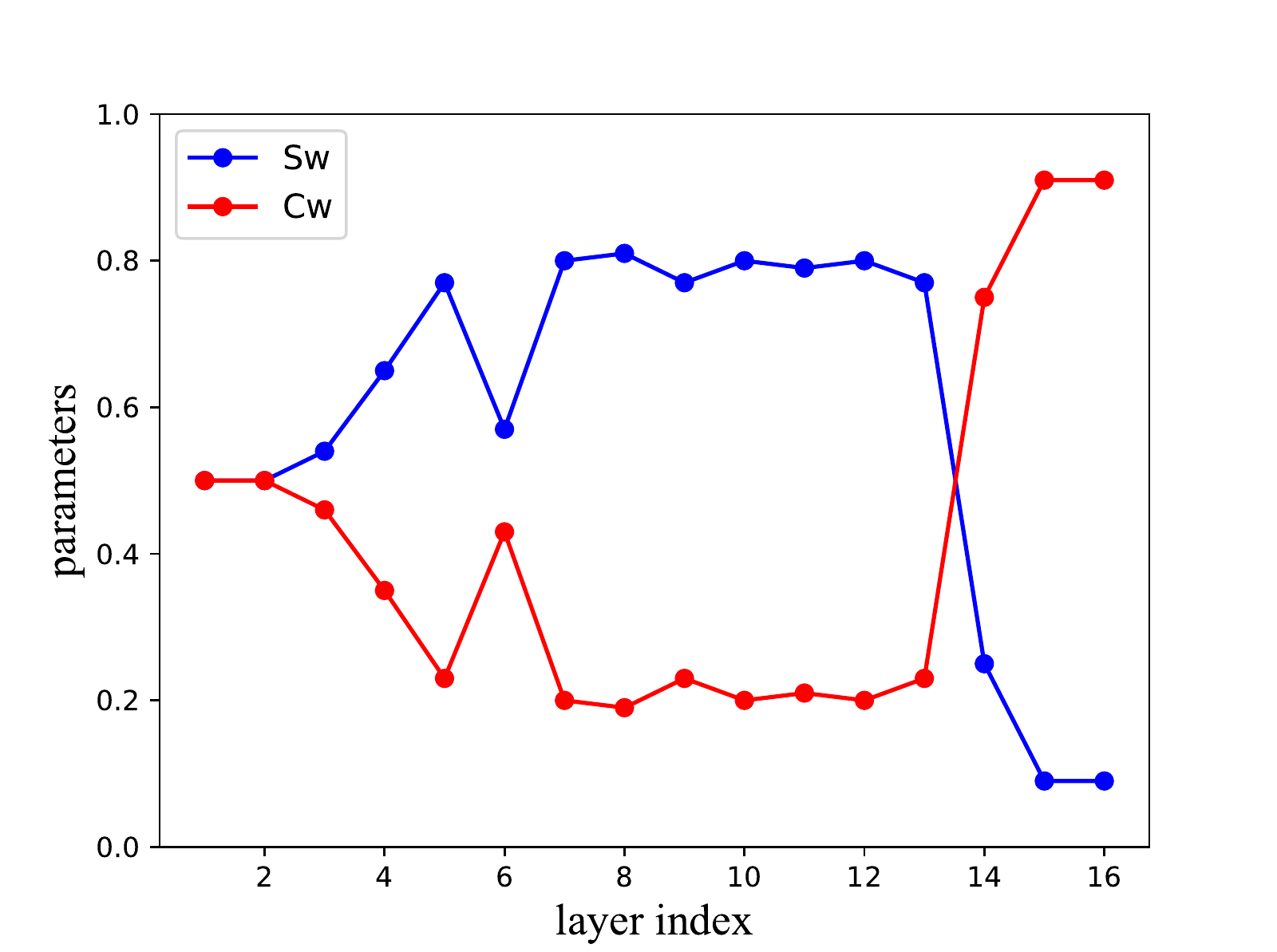}  
\centerline{(b) \small Instance Segmentation}
\end{minipage}  
\begin{minipage}[t]{0.33\linewidth}  
% \centering  

\includegraphics[width=2.4in]{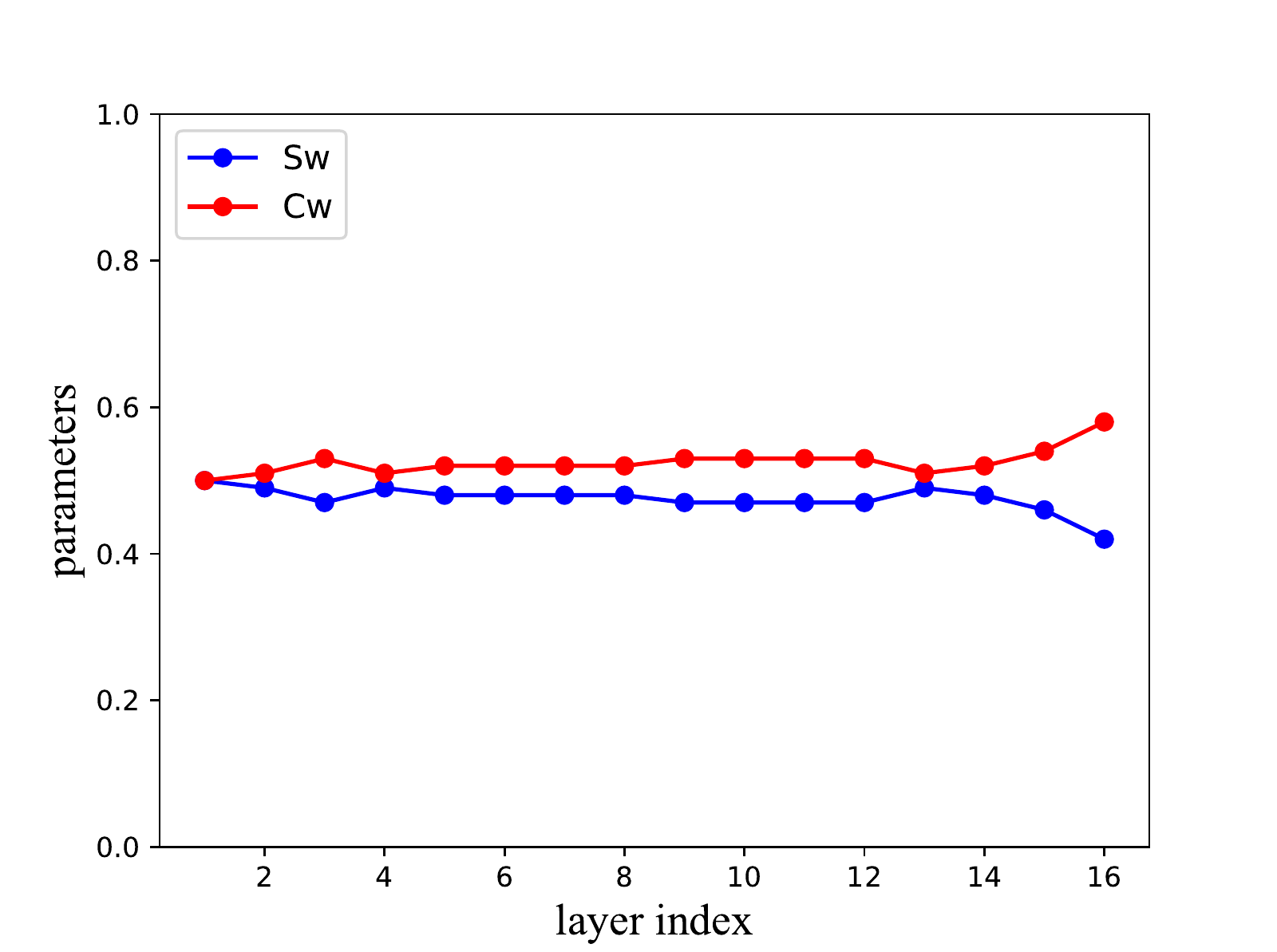}  
\centerline{(c) \small Image Classification}
\end{minipage}  
\vspace{0.5em}
%\caption{Curves of the parameters and the embedded layers} 
\caption{Curves of weight parameters ($S_\text{W}$, $C_\text{W}$) in different tasks with the change of embedding layers. Obviously, the collaborative relationships of these two parameters are different in diverse embedding hierarchies and tasks.}
\label{fig:Figure05}
\end{figure*}
In this section, we evaluate the proposed attention module CAT on MS COCO and Pascal VOC datasets for object detection with Faster-RCNN \cite{article3} and RetinaNet \cite{article4} frameworks.
We also evaluate CAT for instance segmentation with Mask-RCNN \cite{article7}, which is trained on MS COCO, with Cifar-100 dataset for image classification based on four backbones (ResNet50 \cite{article42}, ResNet101 \cite{article42}, MobileNetV2 \cite{article43}, and ShuffleNetV2 \cite{article44}). Specifically, we conduct CAT module on ImageNet for image classification.

\subsection{Object Detection Experiments}

We conduct object detection on Pascal VOC and MS COCO 2017 datasets and deploy all experiments on the Detectron2 platform. In this section, we adopt the average $AP$, $AP_{50}$, and $AP_{75}$ to evaluate the effects of the attention frameworks and use the Faster-RCNN and RetinaNet as detection architectures.

\textbf{Object Detection on Pascal VOC:} Pascal VOC dataset contains 20 categories. 
The training set includes 22136 images for VOC-2007 and VOC-2012 trainval and the test set includes 4952 images for VOC-2007. 
We adopt FasterRCNN as the detection architecture and load the model pretrained on ImageNet as our backbone network. 
In detail, we train the model on 4 Geforce GTX TITAN X GPUs and the batch size is set to 8. 
We take synchronous stochastic gradient descent (SGD) as an optimizer with a momentum of 0.9, weight decay of 0.0001, and total iteration number of 18,000. 
The initial learning rate is set to 0.02 and it drops by a factor of 10 after 12,000 and 16,000 iterations. 

The experimental results are summarized in Table \ref{table01}. The bold and underlined data indicate the optimal and sub-optimal results. 
It shows when applying Faster-RCNN as the basic detector, CAT is superior to the original ResNet by 2.07$\%$ and 1.23$\%$ in terms of $AP$ when the network depth is 50 and 101. Meanwhile, CAT achieves 2.26$\%$ and 1.07$\%$ gains over SENet using ResNet50 and ResNet101. On the other hand, CAT outperforms the original ResNet by 1.04$\%$ in terms of $AP$ when the network depth is 50 and 101 using RetinaNet, and it also achieves better performance than all previous arts in terms of all three measures. Specifically, CAT improves CBAM over 0.72$\%$ and 1.57$\%$ for ResNet50 and ResNet101.
When IoU = 0.75, CAT has a remarkable improvement in AP compared with other IoUs. 

\textbf{Object Detection on MS COCO 2017:} MS COCO 2017 includes 118,287 training images and 5000 validation images with 80 categories. Table \ref{table02} shows the performance of attention modules on MS COCO 2017. $AP_\text{S}$, $AP_\text{M}$, $AP_{\text{L}}$ represent the $AP$ value detected by small targets (area$<$32), medium targets (32$<$area$<$96), and large target (area$>$96). The number of training iterations is set to 90,000. The learning rate drops by a factor of 10 after 70,000 and 80,000 iterations. When applying RetinaNet as the basic detector, CAT outperforms the original ResNet by 1.76$\%$ and 0.97$\%$ in terms of $AP$ when the network depth is 50 and 101. Meanwhile, CAT is superior to SENet and ECANet in terms of all precision measures. Besides, CAT outperforms the original ResNet by 1.72$\%$ and 1.14$\%$ in terms of $AP$ when the network depth is 50 and 101 when using Faster-RCNN. Similarly, we can notice that when CAT detects large objects and IoU=0.75, the corresponding AP obtained is the highest improvement compared to the baseline. This indicates that our method is more appropriate for scenarios that require rigorous object location and size, such as autonomous driving.

\subsection{Image Classification Experiments}

\textbf{Image Classification on Cifar-100:} To verify the classification effectiveness of CAT, we employ ResNet50, ResNet101, MobileNetV2, and ShuffleNetV2 as the base networks. Cifar-100 contains 100 categories and each category consists of 600 images (500 of train set and 100 of test set). We compare our module with SENet \cite{article13}, CBAM \cite{article37}, and ECANet \cite{article15} that are trained on the Cifar-100 training set and measure the Top-1 and Top-5 error on the test set. All programs are trained within 200 epochs on 1 Geforce GTX TITAN X GPU, and the batch size is 128. We use SGD with a momentum of 0.9 and a weight decay of 0.0005. The learning rate is initialized to 0.001 and drops by a factor of 10 every 50 epochs. The results are shown in Table \ref{table07}. The CAT based on ResNet101 achieves 1.62$\%$ gain of Top-1 and 1.84$\%$ of Top-5. Besides, CAT improves CBAM on the accuracy of Top-1, Top-5 over 0.31$\%$, 0.74$\%$ and 0.35$\%$, 0.44$\%$ for Mobilenetv2 and Shufflenetv2. And our CAT favorably improves all the strong baselines with negligible additional
parameters. Such improvements demonstrate the efficiency of CAT. 

% Please add the following required packages to your document preamble:
% \usepackage{multirow}
\begin{table}[!h]
\small
\setlength\tabcolsep{11pt}
\textwidth 8.7cm
\centering
\scalebox{0.9}{\begin{tabular}{lcccc}
\hline
     Methods   & Top-1 & Top-5 & GMac & Params(M) \\ \hline
    ResNet50   &  35.60     &  11.98     &    63.99   &   24.37     \\
    + SENet  &   34.93    &   12.17    &  64.08     &   26.22     \\
    + CBAM   &   36.28    &   12.78    &    64.17   &    26.22    \\
    + ECANet &   \textbf{34.36}    &    \underline{11.86}   &   64.08    &     23.71   \\
    + CAT (Ours) &   \underline{34.78}    &   \textbf{11.35}    &  64.20     &    26.22    \\ \hline
    ResNet101   & 36.61  &  13.78  &  123.57 &  42.49      \\
    + SENet  &  \underline{36.24}     &    \underline{13.16}   &  123.72 & 47.44   \\
    + CBAM   &   37.35    &   13.87    & 123.90   & 47.44  \\
    + ECANet &    36.76   &     13.55  & 123.91 & 47.45   \\
    + CAT (Ours) &  \textbf{34.99}     &   \textbf{11.94}    &  123.91 &  47.45 \\ \hline
   Mobilenetv2   & 48.24    &  19.33   & 2.42    & 2.37    \\
    + SENet  &    48.24   &   19.29    &   2.43    &   2.40    \\
    + CBAM   &   \underline{47.65}    &    19.37   &  2.44    &  2.40    \\
    + ECANet &   48.11    &    \underline{19.07}   &  2.43   & 2.37     \\
    + CAT (Ours) &   \textbf{47.34}    &   \textbf{18.63}    &  2.44   &  2.40     \\ \hline
  Shufflenetv2   &  48.87     &  20.07    &  2.23     &   1.40     \\
  + SENet  &    48.09   &   19.09    &    2.24   &    1.40    \\
  + CBAM   &   48.41    &   \underline{19.08}    &    2.25   &    1.40   \\
  + ECANet &    \underline{48.07}   &   19.62    &    2.24   &    1.36    \\
  + CAT (Ours) &    \textbf{48.06}   &   \textbf{18.64}   &   2.25    &   1.40     \\ \hline
  
\end{tabular}}
\vspace{0.5em}
\caption{Image classification results of different attention methods on Cifar-100. }\label{table07}
\end{table}

\textbf{Image Classification on ImageNet:} To evaluate our CAT module on ImageNet classification, we employ ResNet50 and ResNet101 as backbone models. 
We optimize network parameters by SGD with a weight decay of 1e-4, momentum of 0.9, and mini-batch size of 256. 
We train all models for 100 epochs by setting the initial learning rate to be 0.1, which is decreased by a factor of 10 per 30 epochs. 
We compare our CAT module with SENet, CBAM, and ECANet on different backbones. 
As presented in Table \ref{table14}, backbone networks with our CAT outperform all baselines significantly, demonstrating that CAT generalizes well on various models in large-scale datasets. 
Moreover, CAT achieves the best Top-1 and Top-5 accuracy compared to other modules when the network depth is 50. 
Our CAT is competitive to other modules with negligible additional parameters used when the network depth is 101. 
Based on above results, CAT not only boosts the accuracy of baselines significantly but also favorably improves the performance of other modules. 
CAT also has the potential to boost its performance through further investigating the GEP operator and information fusing technique.

\begin{table}[!h]
\small
\setlength\tabcolsep{10pt}
\textwidth 8.7cm
\centering

\scalebox{0.9}{\begin{tabular}{lcccc}

\hline
                              Methods   & Params(M)    & FLOPs(G)  & Top-1 & Top-5   \\ \hline
% \multirow{1}{*}ResNet-200  &74.45 & 14.10 & 78.20 & 94.00 \\ 
% \multirow{1}{*}Inception-v3   &  25.90  &   5.36   &  77.45 & 93.56 \\ 
% \multirow{1}{*}ResNeXt-101 &  46.66  &   7.53   &  78.80 & 94.40  \\ \hline 
% \multirow{1}{*}DenseNet-264 (k=32) &  31.79  &  5.52    &   77.85 & 93.78  \\ 
% \multirow{1}{*}DenseNet-161 (k=48)  & 27.35  &  7.34    &   77.65 & 93.80   \\ \hline
\multirow{1}{*}ResNet50  & 24.37  &  3.86    &   75.44 & 92.50  \\
\multirow{1}{*} + SENet  & 26.77  &  3.87   &   76.86 & 93.30 \\
\multirow{1}{*} + CBAM  & 26.77  &  3.87    &   77.34 & \underline{93.69} \\
\multirow{1}{*} + ECANet &  24.37  &   3.86   &  \underline{77.48} & 93.68    \\
\multirow{1}{*} + CAT (ours)  & 26.51 &  3.95  &  \textbf{77.99}    &  \textbf{94.14}    \\ \hline
\multirow{1}{*}ResNet101  & 42.49  &  7.34    &   76.62 & 93.12  \\
\multirow{1}{*} + SENet  & 47.01  &  7.35    &   77.65 & 93.81 \\
\multirow{1}{*} + CBAM  & 47.01  &  7.35    &   78.49 & 94.31 \\
\multirow{1}{*} + ECANet  & 42.49 &  7.35  &  \underline{78.65}    &  \textbf{94.34}    \\ 
\multirow{1}{*} + CAT (ours)  & 44.15 &  7.81  &  \textbf{78.74}    &  \underline{94.32}    \\ \hline
                              
\end{tabular}}
\vspace{0.5em}
\caption{Accuracy comparisons with different attention methods on ImageNet.}
\label{table14}
\end{table}

\subsection{Instance Segmentation Experiments}

\begin{table}[!b]
\small
\setlength\tabcolsep{7pt}
\textwidth 8.7cm
\centering

\scalebox{0.9}{\begin{tabular}{lcccccc}
\hline
Methods  & $AP$ & $AP_{50}$ & $AP_{75}$ & $AP_{S}$ & $AP_{M}$ & $AP_{L}$ \\ \hline
ResNet50  &  33.45  &   53.54    &   35.93    &  15.98   &  35.58   &   48.18  \\
+ SENet  &  33.56  &   53.94    &  35.85     & 16.03    &  35.60   &   \underline{48.55}  \\
+ CBAM   & 33.57   &  54.04     & 35.91      & \underline{16.49}     & \underline{36.01}    &  47.85  \\
+ ECANet &  \underline{33.87}  &   \underline{54.30}    &   \underline{35.99}    &  16.01   &   35.54  &  48.24   \\
+ CAT (Ours) &  \textbf{34.22}  &   \textbf{55.06}    &  \textbf{36.53}     & \textbf{16.91}    &  \textbf{36.60}   &   \textbf{48.62}  \\ \hline
ResNet101  &  35.06  &    55.38   &    37.55   &   16.85  &   37.36  &   50.74  \\
+ SENet &  35.13  &   55.55    &    37.77   &  16.62   &  37.52   &  50.97   \\
+ CBAM   &  34.74  &    55.40   &   37.15    &  16.93   &   37.37  &  50.12   \\
+ ECANet &  \underline{35.67}  &   \underline{56.77}    &   \underline{38.24}   &  \textbf{17.92}   &  \textbf{38.65}   &  \underline{51.29}   \\
+ CAT (Ours) &  \textbf{35.86}  &   \textbf{57.02}    &   \textbf{38.45}    &   \underline{17.55}  &   \underline{38.43}  & \textbf{51.59}    \\ \hline

\end{tabular}}
\vspace{0.5em}
\caption{Instance segmentation results of different attention methods on MS COCO val2017 using Mask R-CNN. }\label{table08}
\end{table}

We deploy instance segmentation experiments on MS COCO 2017 with Mask-RCNN as the basic detector, and ResNet50, ResNet101 as backbone models. The implementation details are generally the same as those in object detection, except that we employ 2 Geforce GTX TITAN X GPUs to train the model. As shown in Table \ref{table08}, CAT outperforms the original ResNet by 0.77$\%$ and 0.8$\%$ in terms of $AP$ when the network depth is 50 and 101. We can also aware that CAT has a better response to large objects. For ResNet50 as backbone, our model achieves the best performance in all indicators of segmentation tasks. For ResNet101 as backbone, which is excepted for the accuracy of small and medium targets, our model gets better AP than ECANet. These results prove the effectiveness of our model in the segmentation tasks.

\subsection{Ablation Study}
The ablation study on Pascal-VOC 2012 for object detection with Faster-RCNN verifies the effectiveness of the channel and spatial collaboration relationship proposed in this paper. As shown in Table \ref{table09},

We firstly introduce a GAP to extract the attention information and compare it with methods that use single attention and sequentially arranged attention modules. 
We realize that the sequential method results are in degraded performance due to the uncontrolled interference between the channel and spatial information. 
The simple combination may not have an ideal performance, the result can inhibit the effect on some characteristics and leads to counter-productive since channel and spatial active features in different ways. In addition, to verify the effectiveness of GEP on model performance improvement, we conducted experiments with and without GEP for different combinations of attention.
Our method based on exterior collaboration tackles this problem by dynamically combining attention modules and obtains 41.81$\%$ in terms of $AP$. Moreover, we introduce an additional interior collaboration approach on the previous model for simultaneously combining GEP, GAP, and GMP operators. Its performance in $AP$ rises to 42.61$\%$.

\begin{table}[!h]

\small
\setlength\tabcolsep{14pt}
\textwidth 8cm
\centering
\scalebox{0.78}{\begin{tabular}{llccc}
\hline
 Methods   & \multicolumn{1}{c}{$AP$}    & \multicolumn{1}{c}{$AP_{50}$}   &\multicolumn{1}{c}{$AP_{75}$}   \\ \hline
\multirow{1}{*}ResNet50 (baseline)  &40.06 & 71.13 & 41.61 \\ \hline

\multirow{1}{*}+ spatial   &  40.21  &   71.09   &  41.20 \\ 
\multirow{1}{*}+ spatial w/o GEP &  40.16  &   71.01   &  41.17  \\  \hline

\multirow{1}{*}+ channel &  40.44  &   71.53   &  41.15  \\  
\multirow{1}{*}+ channel w/o GEP &  40.38  &   71.31   &  41.1  \\  \hline

\multirow{1}{*}+ channel + spatial &  36.85  &  67.44    &   36.46  \\ 
\multirow{1}{*}+ channel + spatial  w/o GEP &  36.61  &  67.05    &   36.29  \\ \hline

\multirow{1}{*}+ spatial + channel  & 37.37  &  68.31    &   37.16   \\ 
\multirow{1}{*}+ spatial + channel  w/o GEP & 37.12  &  67.92    &   36.81   \\ \hline

\multirow{1}{*}CAT w/ exterior colla-factors &  41.81  &   72.99   &  43.53    \\
\multirow{1}{*}CAT w/ exterior $\&$ interior colla-factors  &  \textbf{42.61}  &  \textbf{73.71}    &  \textbf{44.87}    \\ \hline

\end{tabular}}
\vspace{0.4em}
\caption{The comparison of different attention methods with GEP and without GEP on Pascal VOC validation 2012. CAT without exterior colla-factors and interior colla-factors respectively refer to the weights apply in exterior collaboration approach for different attention modules and interior collaboration approach for three attention operators.}
\label{table09}

\end{table}

\subsection{Visualization}

Line plots in Fig. \ref{fig:Figure05}, according to priority, illustrate the change of parameters against embedding layers for object detection, instance segmentation, and image classification tasks with ResNet50 as backbone networks. 
The first two plots show similar patterns in low-level layers of the network. 

$S_\text{w}$ values are usually larger than $C_\text{w}$ values due to the fact that low-level feature maps mainly extract various spatial features (such as textures, contours, and edges) to effectively guide the network to learn “where” to focus. 
On the other hand, at high-level layers, $C_\text{w}$ values gradually exceed $S_\text{w}$ values because rich semantic information in high-level feature maps can effectively promote the network's learning of “what” the object is. 

However, for image classification task, $C_\text{w}$ is always higher than $S_\text{w}$ at all layer levels. 

Such phenomenon reveals that classification networks pay more attention to objects’ categories other than their locations. 
In conclusion, dynamical learning the relationship between channel and spatial modules for information fusion can improve the guidance ability of the overall attention framework.

\section{Conclusion}
In this paper, we propose CAT, a novel attention framework and
design it for adaptively learning to collaborate the contributions of
attention modules and pooling operators, respectively. Specifically,
we introduce GEP to complement GAP and GMP for information
extraction and study their traits’ inherent collaboration relationship
for better information fusion. Our CAT is a plug-and-play framework such that no expensive extra cost is required for applying
it in networks. Extensive experiments demonstrate that CAT can
improve networks’ adaptability in various image hierarchies and
tasks (such as classification, object detection, and instance segmentation) compared to previous state-of-the-art attention-based networks.
We integrate our experiments and conclude that our CAT framework
indicates the potential of elastically designed attention mechanisms
in CNNs for computer vision tasks. We will develop our framework
in more architectures to enhance its ability to explore interactive
collaborations of attention modules.

\bibliographystyle{iet}
\bibliography{sample}

\end{document}